\documentclass[runningheads]{llncs}
\usepackage{graphicx}
\usepackage{comment}
\usepackage{amsmath,amssymb}

\usepackage{symbols}
\usepackage{booktabs}       % professional-quality tables
\usepackage{amsfonts}       % blackboard math symbols
\usepackage{nicefrac}       % compact symbols for 1/2, etc.
\usepackage{microtype}      % microtypography
\usepackage[dvipsnames]{xcolor}
\usepackage{algorithm,algorithmic}
\usepackage{multirow}
\usepackage{soul, color}
\usepackage{subcaption}
\usepackage{array}
\usepackage{wrapfig}

\usepackage{float}
\usepackage{xspace}
\usepackage{inconsolata}

\usepackage{cite}
\definecolor{citecolor}{HTML}{2980b9}
\definecolor{linkcolor}{HTML}{c0392b}
\usepackage[pagebackref=true,breaklinks=true,colorlinks,bookmarks=false,citecolor=citecolor,linkcolor=linkcolor]{hyperref}

\belowdisplayskip \abovedisplayskip

\usepackage{letltxmacro}

\LetLtxMacro{\oldsection}{\section}
\renewcommand{\section}[1]{
    \vspace{-0.11in}
    \oldsection{#1}
    \vspace{-0.10in}
}

\LetLtxMacro{\oldsubsection}{\subsection}
\renewcommand{\subsection}[1]{
    \vspace{-0.09in}
    \oldsubsection{#1}
    \vspace{-0.08in}
}

\LetLtxMacro{\oldsubsubsection}{\subsubsection}
\renewcommand{\subsubsection}[1]{
    \vspace{-0.06in}
    \oldsubsubsection{#1}
    \vspace{-0.05in}
}

\newcommand{\Lagr}{\mathcal{L}}
\newcommand{\rgb}{\texttt{RGB}\xspace}
\newcommand{\depth}{\texttt{Depth}\xspace}

\newcommand{\compassgps}{GPS+Compass\xspace}
\newcommand{\xhdr}[1]{\vspace{0pt}\noindent\textbf{#1}\xspace}

\begin{document}
\pagestyle{headings}
\mainmatter
\def\ECCVSubNumber{3023}  % Insert your submission number here

\title{Seeing the Un-Scene: Learning Amodal Semantic Maps for Room Navigation} % Replace with your title

% CAMERA READY SUBMISSION
\titlerunning{Seeing the Un-Scene}

\author{
Medhini Narasimhan\inst{1,2}\thanks{Work done while an intern at Facebook AI Research. Correspondence to \texttt{medhini@berkeley.edu}}~
Erik Wijmans\inst{3,1}~
Xinlei Chen\inst{1}~ \newline{}
Trevor Darrell\inst{2}~
Dhruv Batra\inst{1,3}~
Devi Parikh\inst{1,3}~\newline{}
Amanpreet Singh\inst{1}~
}

\authorrunning{M. Narasimhan et al.}
\institute{\scriptsize{$^1$~Facebook AI Research~$^2$~University of California, Berkeley~$^3$~Georgia Institute of Technology}}

\maketitle

\begin{center}
    \centering
    \includegraphics[width=\linewidth]{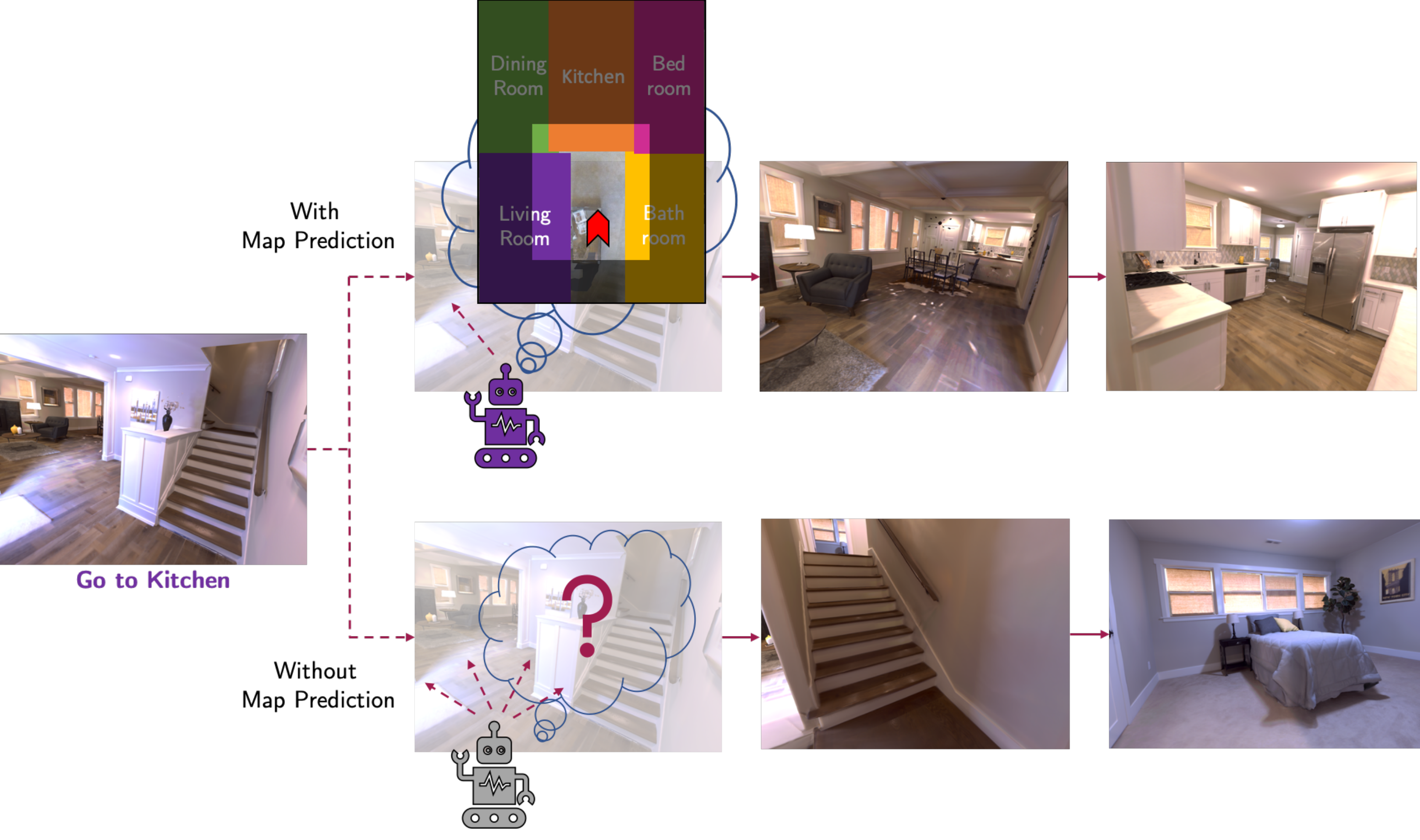}
    \captionof{figure}{
    We train an agent that can navigate to rooms in unseen environments by predicting amodal semantic maps. Here, the agent is placed in the entryway and is asked to navigate to the kitchen. It recognizes a dining room and predicts that the kitchen must be near, even though it has not seen it yet, and heads in that direction. It updates its semantic belief maps as it sees more of the house. Our agent learns to model architectural and stylistic regularities in houses, which helps it find its way in a novel house.
    }
    \label{fig:main}
\end{center}

\begin{abstract}
We introduce a learning-based approach for room navigation using semantic maps. Our proposed architecture learns to predict top-down belief maps of regions that lie beyond the agent’s field of view while modeling architectural and stylistic regularities in houses. First, we train a model to generate amodal semantic top-down maps indicating beliefs of location, size, and shape of rooms by learning the underlying architectural patterns in houses. Next, we use these maps to predict a point that lies in the target room and train a policy to navigate to the point. We empirically demonstrate that by predicting semantic maps, the model learns common correlations found in houses and generalizes to novel environments. We also demonstrate that reducing the task of room navigation to point navigation improves the performance further. 

\end{abstract}

\section{Introduction}
Humans have an uncanny ability to seamlessly navigate in unseen environments by quickly understanding their surroundings. Consider the example in \figref{fig:main}. You're visiting a friend's home for the first time and you want to go to the kitchen. You're in the entryway and you look around to observe your surroundings. You see a bedroom in one direction and the dining room in the other direction. A possible, but tedious solution is to head in a random direction and exhaustively search the space until you end up in the kitchen. Another option, and most probably the one you'd pick, is to walk towards the dining room as you are more likely to find the kitchen near the dining room rather than the bedroom. We believe that there are underlying architectural principles which govern the design of houses and our prior knowledge of house layouts helps us to navigate effectively in new spaces. We also improve our predictions of how the rest of the house is laid out as we walk around and gather more information. The goal of this work is to elicit a similar behaviour in embodied agents by enabling them to predict regions which lie beyond their field of view through learned scene priors. The agent models correlations between the appearance and architectural layout of houses to efficiently navigate in unseen scenes.

Currently, there exist two paradigms for the navigation problem (1) Classical path planning based methods: SLAM-based approaches which first build a geometric map and then use path planning with localization for navigation \cite{hartley2003multiple, thrun2005probabilistic, lavalle2006planning}. (2) Learning-based methods: A policy is learned for a specific task in an environment \cite{zhu2017target, gupta2017cognitive, yang2018visual}. In this work, we introduce a learning-based method that unlike previous approaches, predicts an intermediate representation that captures the model's current belief of the \emph{semantic layout of the house}, beyond the agent's field of view. \cite{wu2019bayesian, yang2018visual} represent priors in the form of knowledge graphs or probabilistic relation graphs which capture room location priors. However, we also want to learn other correlations such as estimating the shape of a room by observing parts of it. We choose to model correlations as semantic maps indicating the location, shape, and size of rooms which lie beyond the agent's field of view. This offers a more flexible representation.

In this work, we develop a novel technique to dynamically generate amodal semantic top-down maps of rooms by learning architectural regularities in houses and to use these predicted maps for room navigation. We define the task of room navigation as navigating to the nearest room of the specified type. For e.g., navigating to the bedroom closest to the starting point. We train an agent using supervision to predict regions on a map that lie beyond its field of view which forces it to develop beliefs about where a room might be present before navigating to it. The agent constantly updates these beliefs as it steps around the environment. The learned beliefs help the agent navigate in novel environments. 

\textbf{Contributions}
(1) We introduce a novel learning-based approach for room navigation via amodal prediction of semantic maps. The agent learns architectural and stylistic regularities in houses to predict regions beyond its field of view.
(2) Through carefully designed ablations, we show that our model trained to predict semantic maps as intermediate representations achieves better performance on unseen environments compared to a baseline which doesn't explicitly generate semantic top-down maps. 
(3) To evaluate our approach, we introduce the room navigation task and dataset in the Habitat platform~\cite{savva2019habitat}.
\section{Related Work}
\label{rel}

\textbf{Navigation in mobile robotics.} 
Conventional solutions to the navigation problem in robotics are comprised of two main steps: (1) mapping the environment and simultaneously localizing the agent in the generated map (2) path planning towards the target using the generated map. Geometric solutions to the mapping problem include (i) structure from motion and (ii) simultaneous localization and mapping (SLAM) \cite{durrant2006simultaneous, bailey2006simultaneous, hartley2003multiple, fuentes2015visual, thrun2005probabilistic, cadena2016past}. Various SLAM algorithms have been developed for different sensory inputs available to the agent. Using the generated map, a path can be computed to the target location via several path planning algorithms \cite{lavalle2006planning}. These approaches fall under the passive SLAM category where a human navigates around the environment beforehand to generate the maps. On the other hand, active SLAM research focuses on dynamically controlling the camera for building spatial representations of the environment. Some works formulate active SLAM as Partially Observable Markov Decision Process and use either Bayesian Optimization~\cite{martinez2009bayesian} or Reinforcement Learning~\cite{kollar2008trajectory} to plan trajectories that lead to accurate maps. \cite{carlone2014active} and \cite{stachniss2005information} use Rao-Blackwellized Particle Filters to choose the set of actions that maximize the information gain and minimize the uncertainty of the predicted maps. 

A less studied yet actively growing area of SLAM research is incorporating semantics into SLAM \cite{walter2013learning, pronobis2012large, bowman2017probabilistic, wang2019computationally}. \cite{wang2019computationally} and \cite{bowman2017probabilistic} use semantic features for improved localization performance and for performing SLAM on dynamic scenes. However, all of the aforementioned SLAM techniques rely on sensor data which is highly susceptible to noise~\cite{chen2019learning}. They also have no mechanism to learn and update semantic beliefs which can be transferred across environments. \cite{aydemir2011plan} combines a classical continual planner with a decision theoretic
planner for active object search. The planner leverages conceptual spatial knowledge in the form of object co-occurrences and semantic place categorisation. On the other hand, our work learns semantic maps of room locations for the task of Room Navigation. \cite{crespo2020semantic} summarizes the various ways of representing semantic information and using it for indoor navigation. The limitations and open challenges in SLAM have been outlined in~\cite{cadena2016past}. This motivates learning based methods for navigation, which we describe next. 

\noindent\textbf{Learning based methods for navigation.} With the motivation of generalizing to novel environments and learning semantic cues, a number of end-to-end learning based approaches have been developed in the recent past~\cite{savinov2018semi, gupta2017cognitive, zhu2017target, mirowski2016learning, fang2019scene}. \cite{savinov2018semi} use a topological graph for navigation and \cite{fang2019scene} propose a memory based policy that uses attention to exploit spatio-temporal dependencies. \cite{mirowski2016learning} jointly learn the goal-driven reinforcement learning problem with auxiliary depth prediction tasks. \cite{wu2018building} introduce the RoomNav task in the House3D simulation platform and train a policy using deep deterministic policy gradient~\cite{lillicrap2015continuous} to solve the same. \cite{chen2019learning} focus on building a task-agnostic exploration policy and demonstrate that this helps for downstream navigation tasks. Most relevant to our work is Cognitive Mapping and Planning~(CMP)~\cite{gupta2017cognitive}. It uses a differentiable mapper to learn a spatial memory that corresponds to an egocentric map of the environment and a differentiable planner that uses this memory alongside the goal to output navigational actions. The maps constructed using this approach only indicate free space and contain no semantic information. On the other hand, we predict semantic top-down maps indicating the location, shape, and size of rooms in the house. Furthermore, unlike CMP where the map corresponds to a top-down view of what the agent is currently seeing, our maps predict regions which lie beyond the agent's field of view by learning architectural regularities in houses. \cite{yang2018visual} uses prior knowledge about spatial and visual relationships between objects represented as a knowledge graph~\cite{krishna2017visual} for the task of semantic navigation. The main disadvantage of such an approach is that the agent cannot modify the graph to learn new priors or update existing beliefs during training. \cite{wu2019bayesian} estimates priors at training time by constructing probabilistic relationship graphs over semantic entities and uses these graphs for planning. However, their graphs don't capture information regarding size and shape of rooms and patterns in houses. Contrary to previous approaches, our work does not rely on pre-constructed maps or knowledge graphs representing priors. We dynamically learn amodal semantic belief maps which model architectural and stylistic regularities in houses. Further, the agent updates its beliefs as it moves around in an environment.

\noindent\textbf{Vision-Language Navigation (VLN).} A different but related task, language guided visual navigation was introduced by~\cite{anderson2018vision}. In VLN, an agent follows language instructions to reach a goal in a home. For \eg, ``Walk up the rest of the stairs, stop at the top of the stairs near the potted plant''. There are multiple works which attempt to solve this problem ~\cite{fried2018speaker,wang2019reinforced,wang2018look}. The room navigation task introduced here is different in that the agent doesn't receive any language based instructions, just the final goal in the form of a room type. In~\cite{anderson2018vision}, the agent “teleports”  from  one  location  to  another  on  sparse pre-computed navigation graph and can never collide with anything. Our Room  Navigation  task  on  the  other  hand,  starts  from  the significantly  more  realistic  setting  as  in  Point  Navigation~\cite{savva2019habitat}  where  the  agent takes the low-level actions such as move-forward (0.25m), turn-left/right (10 degrees) and learns to avoid collisions. In~\cite{anderson2018vision}, paths have an average length of 3.5m whereas the ground truth paths in our task have average length of 125. Overall, compared to VLN the navigation in our task is significantly more challenging. Relative to point navigation where the goal is specified as coordinates, the goal specification in our work is more semantic and closer to language – name of a room. The room navigation task allows us to move towards complex goal specification (e.g., follow an instruction) while keeping the navigation realistic. We'd like to highlight that the methods developed for VLN aren't directly applicable to room navigation as they all rely on intermediate goals in the form of language based instructions which aren't a part of our task specification.

\section{Room Navigation Task}
\label{sec:roomnav}
The agent is spawned at a random starting location and orientation in a \emph{novel} environment and is tasked with navigating to a given target room -- \eg, ``Kitchen.'' If there exist multiple rooms of same type, the agent needs to navigate to the room closest to its starting location. We ensure that the agent never starts in a room of the target room type \ie if the agent is in a bedroom the target room cannot be a bedroom. Similar to~\cite{savva2019habitat}, with each step the agent receives an \rgb image from a single color vision sensor, depth information from the depth sensor and \compassgps that provides the current position and orientation \emph{relative} to the start position. When there's no GPS information available we only need egomotion, which most robotics platforms provide via IMU sensors or odometry. As in~\cite{savva2019habitat, wijmans2019decentralized}, the agent does not have access to any ground truth floor plan map and must navigate using only its sensors. Unlike point navigation~\cite{savva2019habitat}, room navigation is a semantic task, so \compassgps is insufficient to solve the task and only helps in preventing the agent from going around in circles. 

\begin{figure*}[t]
\centering
\includegraphics[width=\textwidth]{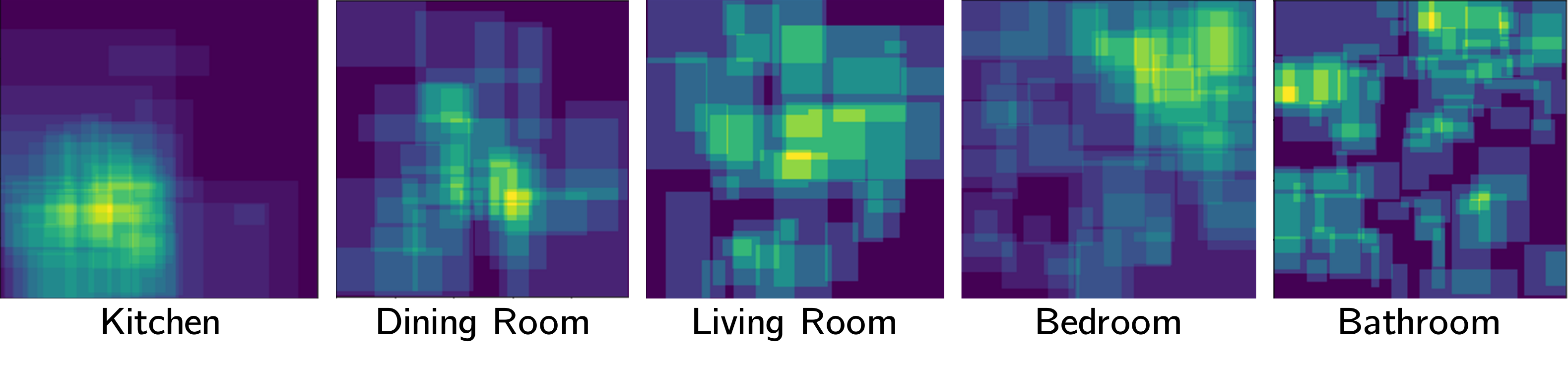}
\caption{\textbf{Architectural patterns in Matterport 3D.} All homes are scaled and aligned such that the kitchen is in the bottom left corner. The maps confirm the existence of structure in house layouts which can be leveraged by an agent trying to navigate to a room in a new environment.} 
\label{fig:priors}
\end{figure*}

\begin{figure*}[t]
\centering
\includegraphics[width=\textwidth]{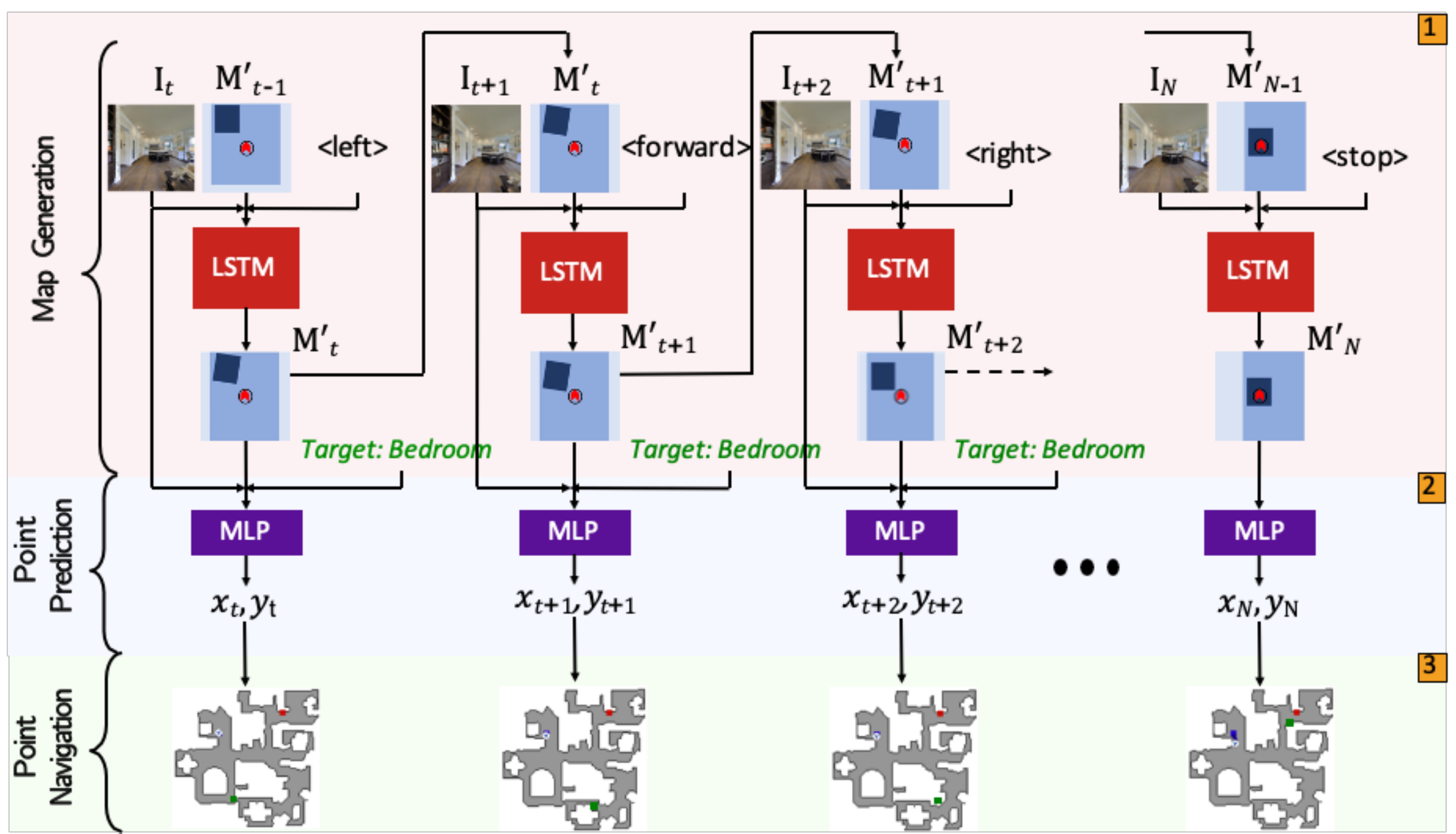}
\caption{\textbf{Room Navigation using Amodal Semantic Maps:} Our room navigation framework consists of 3 components (1) \textbf{Map Generation:} We predict egocentric crops of top down maps indicating rooms in a house. We use a sequence-to-sequence network which takes as input the current image $I_{t}$, the previous semantic map $M'_{t-1},$ and previous $a_{t-1}$ and predicts the current semantic map $M'_{t}$. (2) \textbf{Point Prediction:} We feed the predicted maps ($M'_{t}$) along with $I_{t}$ and target room ID $tr$ to a network to predict a target point $P^\textrm{pred}_{t}=(x_{t}, y_{t})$ that lies in the target room. The predicted and ground truth points are shown in {\color{green}green} and {\color{red}red} respectively on the top down map. (3) \textbf{Point Navigation:} We use a pre-trained point navigation policy to navigate to $P^\textrm{pred}_{t}$. The process is repeated for $N$ steps until the agent calls \texttt{stop} or $N=500$, whichever happens first.} 
\label{fig:overview}
\end{figure*}

\section{Room Navigation using Amodal Semantic Maps}
\label{sec:approach}

\begin{figure*}[t]
\centering
\includegraphics[width=\textwidth]{figs/maps.pdf}
\caption{ \textbf{Egocentric crops of Semantic Top Down Maps.} We show the regions in a $26m \times 26m$ crop of the ground truth semantic maps \wrt agent's location and orientation. Agents are trained to predict these semantic maps, including regions they have not seen yet, based on regions of the house they have seen so far.
Regions colored \protect\includegraphics[height=0.3cm]{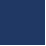} lie inside the specified room. \protect\includegraphics[height=0.3cm]{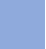} denotes regions inside the house and not in the room and \protect\includegraphics[height=0.3cm]{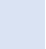} denotes regions outside the house.} 
\label{fig:egocentric_crops}
\end{figure*}

We develop a room navigation framework with an explicit mapping strategy to predict amodal semantic maps by learning underlying correlations in houses, and navigating using these maps. Our approach assumes there are architectural and stylistic regularities in houses which allow us to predict regions of a house which lie beyond our field of view. For instance, if we are in the kitchen we can guess that the dining room is adjacent to it and we would be right in most cases.\footnote{We acknowledge that these regularities likely vary across geographies and cultures.} We first verify the existence of such correlations by scaling and aligning all the homes in the Matterport 3D\cite{chang2017matterport3d} dataset such that the kitchen is in the bottom-left corner. As shown in Fig.~\ref{fig:priors}, we can observe that the concentration of dinning rooms is close to the kitchens and the bedrooms are in the opposite corner, away from the kitchens. We believe there exist similar and more subtle correlations (\eg size of the kitchen could be indicative of the number of bedrooms in a house) which our agent can automatically learn while predicting amodal semantic top down maps of regions. We now provide an overview of our approach followed by a detailed explanation of each of its sub-components.

\noindent\textbf{Overview}. Our room navigation framework is outlined in Fig.~\ref{fig:overview}. The agent is spawned in a random location and is asked to navigate to a specified target room, $tr \in \mathcal{R}$, where $\mathcal{R}$ is the set of all possible target rooms. With each step $t$, the agent receives \rgb image $I_{t}$, \depth $D_{t}$ and \compassgps information. The agent predicts egocentric crops of top down semantic maps indicating the rooms which lie in and beyond its field of view. An example of these maps is shown in~\figref{fig:egocentric_crops}. To generate the maps, the agent uses a sequence to sequence (Seq2Seq) network which takes as input the \rgb Image from the current time-step $I_{t}$, the predicted semantic maps from the previous time step $M^\textrm{pred}_{t-1,r} \forall r \in \mathcal{R}$, and the previous action $a_{t-1}$ and predicts semantic maps $M^\textrm{pred}_{t,r}$ for the current time step. The predicted semantic maps $M^\textrm{pred}_{t,r}$ are fed to a point prediction network which predicts a target point $P^\textrm{pred}_{t}$ lying inside the target room. The agent navigates to the predicted point using a  point navigation policy $\pi_{nav}(P^\textrm{pred}_{t}, D_{t})$. The agent updates its beliefs of the predicted semantic maps and the predicted point as it steps around the environment. The episode is deemed successful if the agent stops inside the target room before 500 steps. 

Next, we describe the three main components of our model architecture: Map Generation, Point Prediction and Point Navigation, as shown in~\figref{fig:overview}.

\subsection{Map Generation}
\label{sec:mapper}
We model the correlations in houses by learning to predict amodal semantic top-down maps. The agent uses the information it has gathered so far to determine where the different rooms in the house are present even before visiting these regions. The maps are crops of top-down maps indicating the type, location, size and shape of rooms and are egocentric to the agent. \figref{fig:egocentric_crops} shows an example of ground truth maps. The maps have 3 classes: (1) Regions which lie outside the house, (2) Regions which lie in the house but outside the target room, (3) Regions which lie in the target room. We hypothesize that the agent will learn architectural regularities in houses and correlations between the \rgb images and layouts in order to predict these maps. We train a map generation network $f_{map}$, to predict egocentric crops of top down semantic maps for each room $r \in \mathcal{R}$. $f_{map}$ consists of a sequence to sequence network $f_{seq}$ and a decoder network $f_{dec}$. At each time step $t$, $f_{seq}$ takes in as input a concatenation of learned representations of the current \rgb frame $f_{i}(I_{t})$, the previous action $f_{act}(a_{t-1})$, and the semantic map of the previous time step $f_{m}(M_{t-1,r})$. $f_{seq}$ outputs a latent representation $h_{t,r}$. $h_{t,r}$ is passed through a parameterized decoder network $f_{dec}$ that resembles the decoder in~\cite{ronneberger2015u}. $f_{dec}$ upsamples the latent representation using multiple transpose convolutions layers to produce output $M^\textrm{pred}_{t,r} \forall r \in \mathcal{R}$. Following~\cite{bengio2015scheduled}, during training, we uniformly choose to set $M_{t,r}^\textrm{input}$ to be $M^\textrm{pred}_{t-1,r}$ (predicted map from previous time step) 50\% of the time and $M^\textrm{GT}_{t-1,r}$ (ground truth map from the previous time step) the rest of the time. At each step, we use the ground truth semantic map $M_{t,r}^\textrm{GT}$ to train $f_{seq}$ and $f_{dec}$ with cross entropy loss, $\Lagr_{map}$. Eqn.~\ref{eq:map_start}-\ref{eq:map_end} describe the exact working of $f_{map}$. The agent continuously generates maps at each step until it calls \texttt{stop} or reaches end of the episode.
%\ie completes 500 steps.

\vspace{-1.50em}
\begingroup
\allowdisplaybreaks
\begin{align}
  \label{eq:map_start}
  \mu&~\sim~\text{Uniform}(0, 1)\\
  M_{t,r}^\textrm{input} &= 
  \begin{cases}
      M_{t-1,r}^\textrm{GT}, \text{if } \mu > 0.5~\text{or t =}~0,\\
      M^\textrm{pred}_{t-1,r}, \text{otherwise }
  \end{cases}\\
  h_{t,r} &= f_{seq}(f_m(M^\textrm{input}_{t}), f_{act}(a_{t-1}), f_i(I_t)) \\
  M^\textrm{pred}_{t,r} &= f_{dec}(h_{t,r}) \\
  \Lagr_{map} &= \sum_{r \in \mathcal{R}} \text{CrossEntropy}(M^\textrm{pred}_{t,r}, M_{t,r}^\textrm{GT})
  \label{eq:map_end}
\end{align}
\endgroup

During inference, we feed a random image as the input semantic map for the first time step and use $M_{t,r}^\textrm{pred}$ as $M_{t,r}^\textrm{input}$ for all consecutive steps.

\subsection{Point Prediction}
\label{sec:point_pred}
The maps predicted by $f_{map}$ are amodal -- the agent predicts regions that it has not seen yet. They are however \emph{crops} -- the agent does not predict the layout of the entire house. These crops are egocentric to the agent and the target room may not always appear in these maps. For \eg, consider~\figref{fig:egocentric_crops}, there exists a bathroom in the house but this region does not appear inside the crop as it does not fall inside the crop \wrt the agent's current location. Inspired by the recent progress in point navigation~\cite{savva2019habitat, wijmans2019decentralized}, we reduce the room navigation problem to point navigation. 
We train a network $f_{point}$ to predict a target point $P^\textrm{pred}_{t}=(x'_{t}, y'_{t})$ that lies in the target room $tr$, at each step $t$ of the agent. Similar to Sec.~\ref{sec:mapper}, we learn representations $f_{i}(I_{t})$ of the \rgb image $I_t$, $f_m(M_{t,r}^\textrm{pred}) \forall r \in \mathcal{R}$ of the predicted semantic maps $M_{t,r}^\textrm{pred}$, and $f_{emb}(tr)$ which is a one-hot embedding of the target room ID $tr$. The predicted semantic map representations for the different rooms are combined using Eq.~\ref{eq:combine_maps}, %
\be
g_{M} = f_m(M_{t,r}^\textrm{pred}) \odot f_{emb}(tr) 
\label{eq:combine_maps}
\ee 
where $\odot$ represents element-wise multiplication. These are then
concatenated with the target room ID $tr$ and fed to a multilayer perceptron (MLP) $f_{point}$ which outputs $P_t^\textrm{pred}$ as described in Eqn.~\ref{eq:b_point}. $f_{point}$ is trained using mean square loss \wrt a ground truth target point in the target room, $P^\textrm{GT}$, as shown in Eqn.~\ref{eq:e_point}. Sec.~\ref{sec:dataset} describes how this point is chosen. 
\begingroup
\allowdisplaybreaks
\begin{align}
\label{eq:b_point}
  P_t^\textrm{pred} = (x'_t, y'_t) &= f_{point}(g_{M}, f_{i}(I_{t}), f_{emb}(tr)) \\
  \Lagr_{point} &= \text{MSELoss}(P^\textrm{pred}_t, P^\textrm{GT}) 
  % a_t &= \pi_{nav}(P_t, I_t, D_t) \\
\label{eq:e_point}
\end{align}
\endgroup

During inference, the agent predicts a point every $k=6$ steps. Once the agent completes $K=60$ steps, the target point is simply fixed and no longer updated. The episode terminates if the agent calls \texttt{stop} or reaches $N=500$ steps.

\subsection{Point Navigation}
\label{sec:point_nav}
At this stage, we have reduced room navigation to point navigation where the agent needs to navigate to the predicted target point $P^\textrm{pred}_{t}$. Following the approach in~\cite{wijmans2019decentralized}, we train a point navigation policy using Proximal Policy Optimization (PPO)~\cite{schulman2017proximal} on the dataset of point navigation episodes in~\cite{savva2019habitat}. The policy, described in Eqn.~\ref{Eq:pointnav}, is parameterized by a $2$-layer LSTM with a $512$-dimensional hidden state. It takes three inputs: the previous action $a_{t-1}$, the predicted target point $P_t^\textrm{pred}$, and an encoding of the \depth input $f_{d}(D_t)$. We only feed \depth input to the  point navigation policy as this was found to work best~\cite{savva2019habitat}. The LSTM's output is used to produce a softmax distribution over the action space and an estimate of the value function.
\be
a_t = \pi_{nav}(P_t, f_{d}(D_t), a_{t-1})
\label{Eq:pointnav}
\ee
\noindent
The agent receives terminal reward $r_T = 2.5$, and shaped reward $r_t (a_t) = -\Delta_{\text{geo\_dist}} - 0.01$, where $\Delta_{\text{geo\_dist}} = d_t - d_{t-1}$ is the change in geodesic distance to the goal by performing action $a_t$. We then use this pre-trained policy, $\pi_{nav}$, to navigate to the predicted point, $P^\textrm{pred}_t$. We also fine-tune $\pi_{nav}$ on the points predicted by our model and this improves the performance. 

To recap, $f_{map}$ generates the semantic map of the space, $f_{point}$ acts as a high-level policy and predicts a point, and the low level point navigation controller $\pi_{nav}$ predicts actions to navigate to this point. 

\subsection{Implementation Details}
\label{sec:implementation}

\noindent\textbf{Map Generation.} 
The image and map representations, $f_{i}(I_{t})\in\mathbb{R}^{256}$ and $f_m(M_{t,r}^\textrm{input})\in\mathbb{R}^{256}$, are obtained by first embedding the input \rgb image $I_{t}$ and semantic map $M_{t}^\textrm{input}$ using a ResNet50~\cite{he2016deep} pre-trained on ImageNet~\cite{deng2009imagenet} followed by a fully connected layer. The action representation $f_{act}(a_{t})\in \{0,1\}^{32}$ is a one-hot embedding of the action $a_{t}$. 

All three are concatenated to form a 544 dimensional vector which is reduced to a vector of size 512 using a linear layer. This is fed through $f_{seq}$, a 2 layer LSTM~\cite{hochreiter1997long} which outputs a hidden state $h_{t,r} \in\mathbb{R}^{512}$. This is passed through $f_{dec}$ which consists of 5 transpose convolutions interleaved with BatchNorm~\cite{ioffe2015batch} and followed by ReLU. 
The map generation network is trained using data collected from shortest path trajectories between source and target points in the house. We experimented with multiple semantic map crop sizes ranging from 20m to 40m and found 26m to work the best. 

\noindent\textbf{Point Prediction.} 
The input \rgb image $I_{t}$ and semantic maps $M_{t,r}^\textrm{input}$ are embedded using a pre-trained ResNet50~\cite{he2016deep}, same as before, but $f_m(M_{t,r}^\textrm{input})$ is now a 32 dimensional vector using a fully-connected layer. The target room embedding $f_{emb}(tr)\in\mathbb{R}^{32}$ is a 32 dimensional one-hot encoding of the target room $tr$. The generated maps for each room type $f_m(M_{t,r}^\textrm{input})$ are multiplied with target embedding $f_{emb}(tr)$ and concatenated to form $g_M\in\mathbb{R}^{160}$. This is concatenated with the \rgb image to form a 416 dimensional vector which is passed through $f_{point}$ to obtain $P^\textrm{pred}_{t}$. We compute a goal vector relative to the agent's current location which is used by the  point navigation module. Similar to map generation, the point prediction network is also trained using data collected from a shortest path trajectories.

\noindent\textbf{Point Navigation Policy.} The \depth encoding $f_d(D_t)$ is based on ResNeXt~\cite{xie2017aggregated} with the number of output channels at every layer reduced by half. As in ~\cite{wijmans2019decentralized}, we replace every BatchNorm layer~\cite{ioffe2015batch} with GroupNorm~\cite{wu2018group} to account for highly correlated inputs seen in on-policy RL. As in ~\cite{wijmans2019decentralized}, we use PPO with Generalized Advantage Estimation (GAE)~\cite{schulman2015high} to train the policy network. We set the discount factor $\gamma$ to $0.99$ and the GAE parameter $\tau$ to $0.95$. Each worker collects (up to) 128 frames of experience from 4 agents running in parallel (all in different environments) and then performs 2 epochs of PPO with 2 mini-batches per epoch. We use Adam~\cite{kingma2014adam} with a learning rate of $2.5 \times 10^{-4}$. We use DD-PPO \cite{wijmans2019decentralized} to train 64 workers on 64 GPUs.

\section{Room Navigation Dataset}
\label{sec:dataset}

\xhdr{Simulator and Datasets.} 
We conduct our experiments in Habitat~\cite{savva2019habitat}, a 3D simulation platform for embodied AI research. We introduce the room navigation task in the Habitat API and create a dataset of room navigation episodes using scenes from Matterport 3D~\cite{chang2017matterport3d}. We use Matterport 3D as it is equipped with room category and boundary annotations and hence is best suited for the task of Room Navigation. It consists of 61 scenes for training, 11 for validation, and 18 for testing. We only use the subset of 90 buildings which are houses and exclude others such as spas as those locations do not have common room categories with the majority of the dataset. We extract a subset of these scenes which contain at least one of the following room types: Bathroom, Bedroom, Dining Room, Kitchen, Living Room on the first floor. We only use the first floor of the house because, (1) In Matterport3d, the bounding boxes of rooms on different floors overlap at times, \eg the box for a room on the first floor often overlaps with the room right above it on the second floor, making it hard to sample points which lie on the same floor, (2) The floors are uneven, making it difficult to distinguish between the different levels of the house. 

Our dataset is comprised of 2.6 million episodes in 32 train houses, 200 episodes in 4 validation houses, and 500 episodes in 10 test houses.

\xhdr{Episode Specification.} An episode starts with the agent being initialized at a starting position and orientation that are sampled at random from all navigable positions of the environment~\cite{savva2019habitat}. The target room is chosen from $R$ if it is present in the house and is navigable from the starting position. We ensure the start position is not in the target room and has a geodesic distance of at least 4m and at most 45m from the target point in the room. During the episode, the agent is allowed to take up to 500 actions. After each action, the agent receives a set of observations from the active sensors. Statistics of the room navigation episodes can be found in the supplementary.

\begin{wrapfigure}{R}{0.5\textwidth}
\centering
\includegraphics[width=0.45\textwidth]{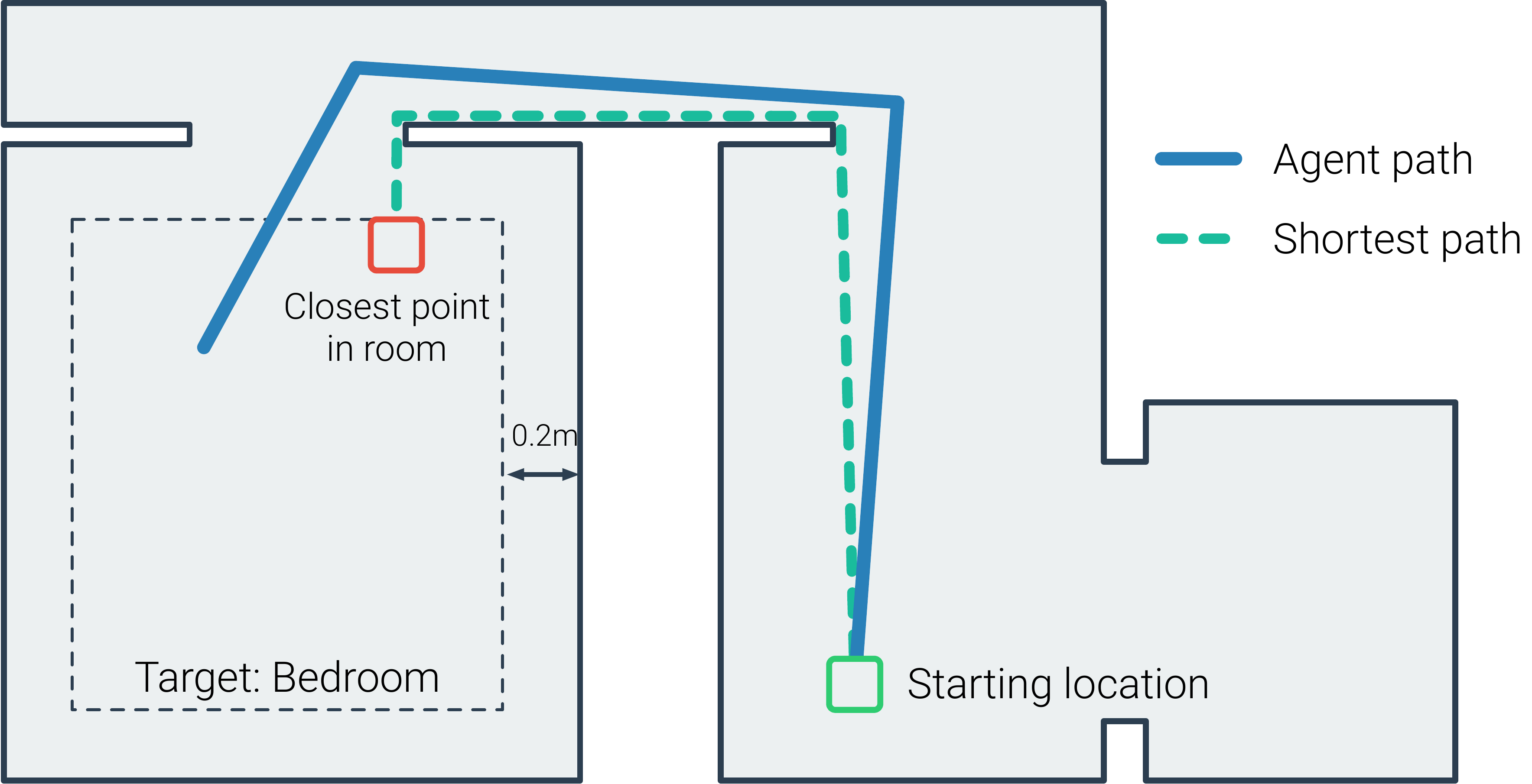}
\caption{\textbf{SPL for Room Navigation.} We compute the geodesic distance from source point to every point inside the room, within 0.2m of the bounds of the room. The point with the smallest geodesic distance is chosen as the \emph{ground truth target point} and the SPL is computed \wrt this point.}
\label{fig:roomspl}
\end{wrapfigure}
\xhdr{Evaluation Metric.} Similar to ~\cite{anderson2018evaluation}, we design two evaluation metrics for room navigation - RoomNav Success weighted by (normalized inverse) Path Length (\emph{\textbf{RoomNav SPL}}) and \emph{\textbf{Success}}. An episode is considered a success if the agent stops 0.2m inside the bounds of the specified target room. We use 0.2m as the room boundaries in Matterport 3D sometimes lie outside the room and this factor ensures the agent has indeed stepped inside the room. As shown in ~\figref{fig:roomspl}, we compute the geodesic distance from the source point to all the navigable points that lie 0.2m within the bounds of the room and choose a ground-truth target point $P^{GT}$ that is closest to the agent's start position, \ie has the shortest geodesic distance. RoomNav SPL is similar to the SPL defined in~\cite{anderson2018evaluation}. Let $S$ indicate `success', $l$ be the length of the shortest geodesic distance between start point and $P^{GT}$ defined above, and $p$ be the length of the agent's path, then 
$
\text{RoomNav SPL} = S \frac{l}{\max(l, p)}
$.
To achieve an SPL of 1, the agent must enter the nearest target room, and call stop when it has stepped 0.2m into the room.

\xhdr{Agent.} As in \cite{savva2019habitat}, the agent is modeled as a cylinder with diameter 0.2m and height 1.5m. The actions and the sensors are the same as in~\cite{savva2019habitat}.

\section{Results}
\label{exp}

We design baselines to evaluate the effectiveness of each component of our proposed room navigation framework and to validate our approach. We also report oracle results using ground-truth annotations to establish an upper-bound on the scores that can be achieved by our model. 

Table~\ref{tab:results} shows the RoomNav SPL and Success scores on the room navigation validation and test sets (for selected baselines).
Our room navigation framework described in Sec.~\ref{sec:approach} achieves an SPL of 0.31 on validation and 0.29 on the test set. Fine-tuning the point navigation policy on points predicted by the point prediction network improves the SPL to 0.35 on validation and 0.33 on test, making this our best performing model.

\vspace{-0.5mm}
\begin{table*}
    \centering
    \setlength{\tabcolsep}{2pt}
    \caption{\small\textbf{RoomNav-SPL for our approach, baselines, and oracle methods on test and validation sets of the Room Nav Dataset.} Our proposed model (Map Generation + Point Prediction + PointNav + Fine-tune) achieves 0.35 RoomNav-SPL and outperforms all other baselines. 
    % \devi{Compared to what of Vanila...? Also, we are comparing to all, right? Why just name one? Just say ``Our proposed model (name) outperforms all baselines. Do you want to add ``Our:'' in front of our approach, ``Baseline:'' in front of all baselines, and ``Oracle:'' in front of approaches that check upper bound? Can remove RoomNav- from the SPL column to get some space.}
    }
    \resizebox{\textwidth}{!}{%
    \begin{tabular}{lc cc cc cc cc}
    \toprule
     \multicolumn{1}{l}{\multirow{2}{*}{\bf RoomNav Model}} && \multicolumn{3}{c}{\bf Validation} && \multicolumn{3}{c}{\bf Test} \\
    \cmidrule{3-5} \cmidrule{7-9}
    && SPL && Success && SPL && Success  \\
    \midrule
    % Gibson-4+ && ResNet50 && 700 && $0.921$ && $0.907$\\
    Baseline: Random && 0.00 && 0.00 && 0.00 && 0.00 \\ \midrule
    Baseline: Vanilla Room Navigation Policy && 0.10 && 0.15 && 0.10 &&  0.11 \\ \midrule
    % Gibson-4+ and MP3D && ResNet50 && 850 && $0.938$ && $0.932$ \\
    Baseline: Map Generation + Room Navigation Policy && 0.16 && 0.17 && - && - \\ 
    Baseline: Point prediction + PointNav && 0.17 && 0.20 && - && - \\ % \hline
    Baseline: Point prediction + PointNav + Fine-tune && 0.21 && 0.23 && - && - \\ \midrule
    Our: Map Generation + Point Prediction + PointNav && 0.31 && 0.35 && - && - \\ % \hline
    Our: Map Generation + Point Prediction + PointNav + Fine-tune && \textbf{0.35} && \textbf{0.38} && \textbf{0.33} && \textbf{0.36} \\ \midrule % \midrule
    Oracle: GT Maps + Room Navigation Policy && 0.54 && 0.56 && - && - \\ % \hline
    Oracle: GT Maps + Point Prediction + PointNav && 0.61 && 0.64 && - && - \\ % \hline
    Oracle: GT Maps + Point Prediction + PointNav + Fine-tune && 0.67 && 0.68 && - && - \\ \midrule
    Oracle: GT Point Selection + PointNav && 0.83 && 0.84 && 0.79 && 0.82 \\
    \bottomrule
    \end{tabular}
    }
    \label{tab:results}
\end{table*}

\xhdr{Vanilla Room Navigation Policy.}
Here we compare to an approach that does not use semantic maps to model correlations and does not use point navigation. We ablate both the map prediction and point generation components by training room navigation policy from scratch using PPO, similar to the point navigation policy in Sec.~\ref{sec:point_nav}. Instead of a target co-ordinates relative to current state, it takes in the target room ID as input. The agent receives terminal reward $r_T = 2.5~\text{RoomNav-SPL}$, and shaped reward $r_t (a_t) = -\Delta_{\text{geo\_dist}} - 0.01$, where $\Delta_{\text{geo\_dist}} = d_t - d_{t-1}$ is the change in geodesic distance to the target room by performing action $a_t$. 
The SPL using this baseline is 0.10 on validation and 0.10 on test, significantly worse compared to our approach (SPL 0.35). \textbf{\emph{This reinforces the effectiveness of our model, specifically the need to generate maps and use point navigation.}} Note that this approach mimics an approach that tries to solve room navigation via vanilla (``brute force'') reinforcement learning.

\xhdr{Vanilla Room Navigation Policy with Map Generation.}
We ablate the point prediction model in Sec.~\ref{sec:roomnav} and train a room navigation policy to navigate to the target room using \rgb images and semantic maps. We use the map generator to predict semantic maps for each room type. We then train a policy to navigate to the target room. The policy is similar to the room navigation policy described above and takes four inputs: the previous action, the target room represented as an ID, the predicted semantic maps for all rooms embedded as in Eq.~\ref{eq:combine_maps} and the \depth encoding. It is trained the same way as the room navigation policy.

This baseline achieves an SPL of 0.16, which is worse by a large margin of ~0.2 when compared to our best performing model (room navigation using Map Generation + Point Prediction). \emph{\textbf{The improved performance of our best method emphasizes the significance of the point prediction and point navigation modules in our best performing model.}}

 \begin{table}[t]
    \centering
\caption{Ablation study of our mapping model and point prediction model.}
\begin{subtable}[t]{0.625\linewidth}
    {
    \centering
    \caption{\small{\textbf{Map Generation Models Performance.} Map Generation is a three-way classification problem. Class-1 consists of regions lying outside the house, Class-2 is regions lying inside the house but not in the room and Class-3 is regions lying in the room.}}
    \setlength{\tabcolsep}{1pt}

    \resizebox{\textwidth}{!}{%
    \begin{tabular}{lc cc cc cc cc cc}
    \toprule
    \multicolumn{1}{c}{\multirow{2}{*}{\bf Mapper}} && 
    \multicolumn{1}{c}{\multirow{2}{*}{\bf mIoU}} && 
    \multicolumn{7}{c}{\bf Pixel Accuracy \%} \\
    \cmidrule{5-11}
    && && Class-1 && Class-2 && Class-3 && Avg \\
    \midrule
    CNN && 25.66 && 31.84 && 36.97 && 5.63 && 24.81 \\ 
    LSTM(no maps) && 32.92 && 41.44 && 48.92 && 10.40 && 33.59\\
    LSTM(ours) && 41.45 && 56.13 && 60.92 && 13.13 && 43.39\\ 
    \bottomrule
    \end{tabular}
    }
    \label{tab:mapper}
    }
    %\end{minipage}
\end{subtable}
\hspace{0.0125\linewidth}
\begin{subtable}[t]{0.325\linewidth}
     {
     \centering
     \caption{\small{\textbf{Point Prediction error using Predicted Maps and Ground Truth Maps.} With ground truth maps, the prediction error is lower, suggesting scope for improvements.}}
     \setlength{\tabcolsep}{1pt}
    \resizebox{\textwidth}{!}{%
     \begin{tabular}{l c}\toprule
         {} & {\bf Error \%} \\
         \midrule
         Pred Maps + MLP & 39.13 \\ 
         GT Maps + MLP & 22.91 \\\bottomrule
     \end{tabular}
    }
     \label{table:point_prediction}
    }
\end{subtable}
\end{table}

\xhdr{Point Prediction and Point Navigation Policy.}
We perform room navigation using only the high-level point prediction network and the low-level point navigation controller.
We ablate the map generation module and train a modified version of the point prediction network without maps as input. Similar to $f_{point}$ in  Eq.~\ref{eq:b_point}, it generates $P_t = (x_t, y_t)$ but by using only the image representation $f_i(I_t)$ and target room embedding $f_{emb}(tr)$. The agent then navigates to $P_t$ using the pre-trained point navigation policy as in Sec.~\ref{sec:point_nav}. 

This method achieves an SPL of 0.17 when the policy is trained from scratch and an SPL of 0.21 when the policy is fine tuned with points predicted by the point prediction network. Our best model surpasses this by a large margin of $\sim$0.15, which \emph{\textbf{shows the advantage of using supervision to learn amodal semantic maps that capture correlations.}} It also indicates the effectiveness of our map generation network. Since the environments in validation are different from train, we can also conclude that predicting semantic maps allows for better generalization to unseen environments as the RoomNav SPL is a direct indicator of how ``quickly'' an agent can reach a target room.

\xhdr{Using Ground Truth (GT) Maps.} To get a better sense of how well our models can do if we had perfect map generation, we train a few of our baselines with ground truth maps instead of generated maps and report results in Table~\ref{tab:results}. With GT maps, Vanilla Room Navigation Policy with Map Generation achieves an SPL of 0.54. Adding GT maps to our best model, with and without fine-tuned point navigation policy we achieve SPL of 0.61 and 0.67 respectively. This suggests that there is still a large room for improvements in the Map Prediction module to perform room navigation more effectively. Table~\ref{table:point_prediction} reports the prediction error of point prediction model when using generated maps and ground truth maps. 

\xhdr{Random.} We evaluate using a random agent that takes action randomly among
\texttt{move\_forward}, \texttt{turn\_left}, and \texttt{turn\_right} with uniform distribution. It calls \texttt{stop} after 60 steps, which is the average length of an episode in our dataset. This achieves a RoomNav SPL of 0 on both test and validation, which implies that our task is difficult and cannot be solved by random walks.

\xhdr{Using GT Point Selection} We use the pre-trained point navigation policy defined in Sec.~\ref{sec:point_nav} to navigate to the ground truth target points $P^\textrm{GT}$ in the target room defined in Sec.~\ref{sec:dataset}. This achieves an SPL of 0.82 and 0.79 on validation and test respectively. It provides an ``upper-bound'' on the performance that can be achieved by the room navigation policy, as this indicates the maximum RoomNav SPL that can be achieved by the framework in Sec.~\ref{sec:approach} if the error on point prediction were 0. These numbers are comparable to the SPL values for point navigation on the Matterport-3D dataset in ~\cite{savva2019habitat}, thus indicating our episodes are at least as difficult as the point navigation episodes in~\cite{savva2019habitat}.

\xhdr{Map Generation Ablations.} We also experimented with different semantic map generation models. The results in Table~\ref{tab:mapper} show that the LSTM map generation model described in Sec.~\ref{sec:approach} performs best with a mean Intersection-over-Union (mIoU) of 41.45 and an Average Pixel Accuracy of 43.39\%. The CNN only approach predicts a semantic map from each \rgb image without maintaining a memory of the previous maps. This performs poorly and has a mIoU of 25.66 and an Average Pixel Accuracy of 24.81\%. We train another LSTM model which doesn't use the semantic maps as input at each time step. This has a mIoU of 32.93 and Average Pixel Accuracy of 33.59. 

\xhdr{Trajectory Videos.} Qualitative results of our model can be found \href{https://youtu.be/5rjvw8dxUw4}{here}. The first image in the first row shows the \rgb input. The second and third maps in the first row show the location of the agent in allocentric and egocentric views respectively. The last figure on the first row shows two dots, red indicating the ground truth target point in the target room and green showing the predicted point. When only one dot is visible it indicates the predicted and ground-truth points overlap. There are 5 ground-truth semantic maps for each of the 5 room types we consider. The labels at the bottom indicate the room type being predicted. The second row shows the ground truth semantic maps indicating the location of the rooms in the house. The third row shows the maps predicted by our agent. The target room is mentioned at the very bottom, in this case, ``Dining Room". As seen in the video, the model dynamically updates the semantic belief maps and predicts the target point with high precision. The agent is able to detect the room its currently present in and also develop a belief of where other rooms lie. The RoomNav-SPL for this episode is 1.0 as the agent successfully reaches the target room following the shortest path. Additional videos can be found \href{https://www.youtube.com/playlist?list=PLzNOWIcHk2AsUW35ECtPCvkzyi4NsNrHD}{here}.

\section{Conclusion}

In this work, we proposed a novel learning-based approach for Room Navigation which models architectural and stylistic regularities in houses. Our approach consists of predicting the top down belief maps containing room semantics beyond the field of view of the agent, finding a point in the specified target room, and navigating to that point using a point navigation policy. Our model's improved performance (SPL) compared to the baselines confirms that learning to generate amodal semantic belief maps of room layouts improves room navigation performance in unseen environments. Our results using ground truth maps indicate that there is a large scope for improvement in room navigation performance by improving the intermediate map prediction step. We will make our code and dataset publicly available. 

\section{Acknowledgements}

We thank Abhishek Kadian, Oleksandr Maksymets, and Manolis Savva for their help with Habitat, and Arun Mallya and Alexander Sax for feedback on the manuscript. The Georgia Tech effort was supported in part by NSF, AFRL, DARPA, ONR YIPs, ARO PECASE, Amazon.  Prof. Darrell’s group was supported in part by DoD, NSF, BAIR, and BDD. The views and conclusions contained herein are those of the authors and should not be interpreted as necessarily representing the official policies or endorsements, either expressed or implied, of the U.S. Government, or any sponsor.

%\clearpage 

\LetLtxMacro{\section}{\oldsection}
{\small
\bibliographystyle{splncs04}
\bibliography{ref}
}

\end{document}